# Three-dimensional attention Transformer for state evaluation in real-time strategy games

Yanqing Ye, Weilong Yang, Kai Qiu, Jie Zhang

*Abstract*—Situation assessment in Real-Time Strategy (RTS) games is crucial for understanding decision-making in complex adversarial environments. However, existing methods remain limited in processing multi-dimensional feature information and temporal dependencies. Here we propose a tri-dimensional Space-Time-Feature Transformer (TSTF Transformer) architecture, which efficiently models battlefield situations through three independent but cascaded modules: spatial attention, temporal attention, and feature attention. On a dataset comprising 3,150 adversarial experiments, the 8-layer TSTF Transformer demonstrates superior performance: achieving 58.7% accuracy in the early game (~4% progress), significantly outperforming the conventional Timesformer's 41.8%; reaching 97.6% accuracy in the mid-game (~40% progress) while maintaining low performance variation (standard deviation 0.114). Meanwhile, this architecture requires fewer parameters (4.75M) compared to the baseline model (5.54M). Our study not only provides new insights into situation assessment in RTS games but also presents an innovative paradigm for Transformer-based multi-dimensional temporal modeling.

## I. INTRODUCTION

REAL time strategy (RTS) games represent one of the most challenging domains for artificial intelligence research, characterized by enormous state and action spaces that dwarf even those of traditional board games like Go [1]. In these complex war simulation environments, players must simultaneously manage economies, control dozens of units, and make strategic decisions under severe time constraints [2]. Despite significant advances in game AI, particularly demonstrated by AlphaGo's triumph over human champions [1], developing strong AI systems for RTS games remains an elusive goal, with top human players still consistently outperforming the best artificial agents [3].

A fundamental challenge in RTS game AI lies in the development of accurate state evaluation functions - methods that can assess the relative advantages of players in any given game state. Traditional approaches have relied primarily on material-based evaluations that consider unit counts, resources, and basic combat metrics [3, 4]. While such methods capture raw military and economic strength, they often fail to account for crucial spatial relationships between units and terrain features that experienced human players naturally recognize and exploit [5]. More sophisticated evaluation techniques based on combat simulation [6] and Lanchester attrition laws [7-10] have shown promise but remain limited in scope, focusing primarily on military aspects while neglecting other strategic elements.

Recent breakthroughs in deep learning, particularly in computer vision [11] and game playing [1], suggest a promising new direction. Convolutional neural networks (CNNs) have demonstrated remarkable ability to learn complex spatial patterns and hierarchical features directly from raw input data [12]. Stanescu et al. [5] investigate the potential of deep CNNs to revolutionize RTS game state evaluation by learning to recognize complex tactical and strategic patterns. Unlike previous approaches that rely on hand-crafted features or focus solely on combat outcomes [13, 14], CNNs can potentially learn to identify and evaluate subtle positional advantages, strategic resource control, and tactical opportunities directly from raw game state representations [5]. Initial work has shown that CNN-based evaluation functions, despite being computationally more intensive, can significantly outperform traditional methods when incorporated into state-of-the-art search algorithms [5, 15, 16]. Their success in Go, where they enabled both accurate position evaluation and intelligent action selection, raises the intriguing possibility of applying similar techniques to the more complex domain of RTS games [5].

Recent years have witnessed the remarkable success of Transformer architecture through its self-attention mechanism in natural language processing and computer vision [17-19]. The key advantages of Transformer include: (1) fewer inductive biases allowing richer feature representation learning from large-scale data [20], (2) direct modeling of long-range dependencies through self-attention rather than stacking multiple convolutional layers to expand receptive fields [21], and (3) a unified token-based processing paradigm for multi-modal inputs [22]. These characteristics make Transformer particularly suitable for handling the complex state spaces and This work was supported by the National Natural Science Foundation of China under Grant No.62103438. *(Corresponding author: Jie Zhang).*

Yanqing Ye is with the Academy of Military Science, Beijing, 100000 China. (e-mail: yeyanqing09@alumni.nudt.edu.cn).
Weilong Yang is with the Academy of Military Science, Beijing, 100000 China.(Co-first author, e-mail: yangweilong09@nudt.edu.cn).
Kai Qiu is with the Academy of Military Science, Beijing, 100000 China. (e-mail: q1987k@163.com).
Jie Zhang is with the Academy of Military Science, Beijing, 100000 China. (e-mail:zhangjie19@alumni.nudt.edu.cn).

long-term dependencies in RTS games.

In video understanding, TimeSformer [23] has achieved remarkable success by extending Transformer's self-attention mechanism to the spatiotemporal domain. Its key innovation lies in proposing "divided attention", which computes temporal and spatial attention separately to maintain modeling effectiveness while improving computational efficiency [24]. With this design, TimeSformer has achieved state-of-the-art performance on multiple video understanding benchmarks while requiring lower training and inference costs [25, 26]. For RTS game state evaluation tasks that require simultaneous understanding of spatial relationships and temporal evolution, TimeSformer's architecture offers inherent advantages over traditional convolutional approaches [5, 27].

However, the conventional TimeSformer architecture is primarily optimized for video classification tasks, with temporal and spatial attention mechanisms mainly processing RGB pixel features [23]. In contrast, each grid position in RTS games contains richer feature information, including unit types, health points, ownership, and resources (neutral or occupied). These features exhibit complex interdependencies, and directly applying the TimeSformer architecture may result in loss of important feature relationship information. Therefore, this study proposes incorporating a Feature Attention (FA) module after spatiotemporal attention to model interactions between different feature dimensions. After modeling temporal and spatial dependencies, this module further processes the associations between feature planes to better capture the intrinsic structure of RTS game states [28]. This multi-dimensional attention mechanism design shows promise in enhancing the model's understanding and evaluation of game situations.

Here, we present a novel channel attention enhanced TimeSformer architecture for RTS game state evaluation that combines the strengths of spatiotemporal attention with explicit feature relationship modeling, demonstrating superior performance in capturing the complex dynamics of RTS games while maintaining computational efficiency.

## II. Materials and Methods

### A. RTS Experimental Design

We developed a comprehensive AI algorithm system for RTS games comprising 10 distinct strategies. These algorithms span from basic random strategies to sophisticated economic-military integrated strategies, enabling simulation of diverse gameplay pat-terns. Based on their characteristics, these algorithms are categorized into three main types: baseline strategies, tactics-oriented strategies, and integrated strategies.

RandomBiasedAI serves as the baseline testing strategy, implementing a biased random decision mechanism. During each decision cycle, this strategy randomly selects potential actions based on predetermined probability distributions, while adjusting se-lection probabilities through bias terms. This design maintains strategic randomness while avoiding potentially irrational behaviors associated with purely random selection.

Tactics-oriented strategies encompass four distinct rush types: WorkerRush, RangedRush, LightRush, and HeavyRush.

Specifically, WorkerRush focuses on rapid worker unit training, launching ear-ly-game attacks through numerical advantage; RangedRush prioritizes ranged combat units, achieving safe offensive positions through range advantages; LightRush empha-sizes light unit mobility, gaining battlefield advantages through speed and flexibility; HeavyRush centers on heavy units, which, despite slower development, provides signif-icant late-game advantages by maintaining continuous heavy unit production alongside economic development, though vulnerable to early harassment strategies like Worker-Rush.

Integrated strategies incorporate more complex decision-making mechanisms. EconomyRush prioritizes economic development, maximizing economic benefits while maintaining basic defense through precise resource collection and allocation algorithms; EconomyMilitaryRush combines economic development with military expansion, seek-ing balance between the two; CRush_V1 and CRush_V2 represent two distinct versions of early-game offensive strategies; PuppetNoPlan employs a reactive decision-making mechanism, responding to immediate battlefield situations rather than following prede-termined plans.

We employed a round-robin tournament system for adversarial testing. Specifically, 10 AI algorithms were paired for head-to-head competitions, with each pair engaging in 70 rounds of matches. To eliminate systematic bias potentially arising from turn order, opponents switched positions after 35 rounds, resulting in a total of $C_{10}^2 \times 70 = 3150$ matches. Experiments were conducted on a 16×16 map (Figure 1). To prevent infinite game loops, we implemented a 1000-round maximum game limit, after which the winner was determined by the number of surviving units.

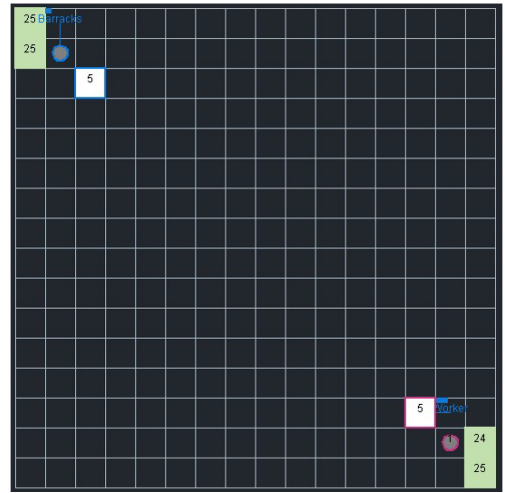

**Figure 1.** 16×16 microRTS map. Green squares represent neutral resources, blue-framed gray squares indicate blue player's barracks, blue-framed orange circles represent blue player's light combat units, blue (red) framed white squares represent blue (red) player's bases, and blue (red) dots represent blue (red) player's worker units.

For dataset partitioning, we adopted a standard ratio of 10:5:2.5 for training:test:validation sets, corresponding to 1,800, 900, and 450 game instances, respectively. The training set comprised 40 rounds (20 rounds each with switched positions) of round-robin matches between 10 neural networks. The validation set included 10 rounds per pairing (5 rounds each with switched positions), while the test set contained 20 rounds per pairing (10 rounds each with switched positions) for final model performance evaluation. This partitioning scheme ensures sufficient training data while maintaining adequate independent test data for model evaluation. Additionally, the validation set size enables effective model selection and hyperparameter optimization.

Tournament results are stored as "environmental state abstract feature matrices". Initially, situational information is collected from the simulation environment as model input. The input data comprises game state recordings at different timestamps. Each entity's position is represented within a 16×16 feature plane representing the map, corresponding to initial RTS game inputs and containing information for all positions in the RTS environment, including entity type, current health value, faction affiliation, neutral resources, and faction resources. Using one-hot encoding, all entity integer attributes (e.g., entity health) are divided into k distinct 16×16 binary planes, ultimately abstracting each temporal state into a 16×16×5 matrix. All planar feature information is shown in TABLE I.

This matrix encompasses five critical dimensions: entity type, health value, faction affiliation, neutral resources, and faction resources. Each dimension employs a specific encoding scheme, including: 7 entity type categories; 11 health levels; binary encoding (1 or 2) for faction affiliation; 26 discrete values (0-25) representing neutral resource quantities, including fixed resource points and resources in transit; and 26 discrete values (0-25) indicating each faction's accumulated resources.

TABLE I
INPUT FEATURE PLANES FOR MULTI-SCALE CONVOLUTIONAL NEURAL NETWORK

| Feature | Number of Types | Description |
|---|---|---|
| Entity Type | 7 | Types 1-7 correspond to base, barracks, resource, worker, light unit, heavy unit, ranged unit |
| Health Value | 11 | Values 0-10, initial health: base 10, barracks 4, resource 1, worker 1, light unit 4, heavy unit 8, ranged unit 1, empty space 0 |
| Faction | 2 | 1 or 2 |
| Neutral Resources | 26 | 0-25, resources in worker transport considered neutral |
| Faction Resources | 26 | 0-25 |

To further capture dynamic state changes during simulation, we extract environmental state abstract feature matrices from n different timestamps and concatenate them into a 16×16×5×n matrix, effectively transforming the input data into 5×n channels. Since n different game states correspond to the same match outcome, this operation not only enriches state feature data but also incorporates temporal dynamic changes into the win/loss prediction process. After determining the input data, based on the neural network designed, we can predict the effect of current planned actions, i.e., the situation assessment results.

Experiments were conducted on a system equipped with an Intel i7 14700K processor and 64GB RAM, running Java version 22.0.1. CPU time per round was constrained to 100 milliseconds.

The microRTS platform employed in our experiments is a well-established framework for RTS algorithm research and development, widely adopted in RTS studies. This platform features a streamlined architecture, multiple game modes, and flexible APIs, enabling researchers to rapidly prototype and experiment while evaluating algorithm performance through adversarial encounters with other AI agents.

*B. Neural Network Architecture*

Given an RTS game situation sequence $X$ as model input, belonging to a five-dimensional real tensor space $X \in \mathbb{R}^{\{B \times T \times C \times H \times W\}}$, where $B$ represents batch size, $T = 500$ denotes discrete time steps, $C = 5$ corresponds to the number of multi-channel feature planes, and $H = W = 16$ represent spatial dimensions of the game map. Each feature plane independently encodes key game state information: entity type matrix $M_{type} \in \mathbb{R}^{H \times W}$ with values in [1,7] representing 7 basic unit types, health value matrix $M_{health} \in \mathbb{R}^{H \times W}$ with values in [0,10] indicating unit survival status, faction matrix $M_{faction} \in \{1,2\}^{H \times W}$ denoting unit affiliation, neutral resource matrix $M_{neutral} \in [0,25]^{H \times W}$ and faction resource matrix $M_{resource} \in [0,25]^{H \times W}$ representing resource distribution states. This multi-channel feature design constructs a complete representation space for game situations.

To enhance computational efficiency and capture local spatial patterns, we partition the input situation map into non-overlapping 4×4 blocks. Each block is mapped to a D-dimensional embedding space ($D = 155$) through a learnable linear transformation matrix $E$, while incorporating positional encoding to preserve spatial and temporal information. This process is represented as:

$$z_{p,t}^0 = Ex_{p,t} + e_{p,t}^{pos} \quad (1)$$

where $z_{p,t}^0 \in \mathbb{R}^D$ is the initial feature representation for position $p$ and time step $t$, $Ex_{p,t}$ is the feature embedding obtained through linear transformation of input block $x_{p,t} \in \mathbb{R}^{C \times 16}$ using learnable matrix $E \in \mathbb{R}^{D \times (C \times 16)}$, and $e_{p,t}^{pos} \in \mathbb{R}^D$ is a learnable positional embedding vector encoding spatiotemporal information in the input sequence. Here, $p \in \{1,2,...,N\}$ denotes spatial position index $\left(N = \frac{H \times W}{16} = 16\right)$, $t \in \{1,2,...,T\}$ represents time step index. Notably, a special token vector $z_{cls}^0 \in \mathbb{R}^D$ is inserted at the sequence start, which aggregates sequence information for situation assessment in the network's final layer. Through this preprocessing, the original input is transformed into a feature sequence with shape $[B, (T \times N + 1), D]$.

We propose a tri-dimensional attention Transformer network comprising $L = 12$ sequentially stacked encoder layers. In each encoder layer $l \in \{1,2,...,L\}$, attention computation is decomposed into three independent but sequentially cascaded submodules: spatial attention (SA), temporal attention (TA), and feature attention (FA). Each encoder layer receives output features $z^{l-1} \in \mathbb{R}^{\{B \times (T \times N+1) \times D\}}$ from the previous layer as input, where the first encoder layer ($l = 1$) takes patch embedding sequence $z^0$ as input.

The SA module initially models spatial dependencies within each time step. For input features $z^l$ of layer l, spatial attention first computes query, key, and value matrices in the spatial dimension:

$$Q_s = W_Q^s z^l + b_Q^s \quad (2)$$
$$K_s = W_K^s z^l + b_K^s \quad (3)$$
$$V_s = W_V^s z^l + b_V^s \quad (4)$$

where $W_Q^s, W_K^s, W_V^s \in \mathbb{R}^{D \times D}$ are learnable weight matrices and $b_Q^s, b_K^s, b_V^s \in \mathbb{R}^D$ are corresponding bias vectors. Spatial attention weights are then computed through scaled dot-product attention:

$$SA(z^l) = softmax\left(\frac{Q_s K_s^T}{\sqrt{d_h}}\right) V_s. \quad (5)$$

Here, $d_h = \frac{D}{h} = 31$ represents the dimension per attention head, with $h = 5$ attention heads. During computation, input is reshaped to $[(B \times T) \times N \times D]$, enabling direct attention calculation across N spatial positions within each time step. This multi-head design allows parallel capture of spatial dependencies in different subspaces.

Building on spatial relationship modeling, the TA module captures temporal patterns in battle evolution. This module computes query, key, and value matrices in the temporal dimension based on spatial attention output:

$$Q_t = W_Q^t SA(z^l) + b_Q^t \quad (6)$$
$$K_t = W_K^t SA(z^l) + b_K^t \quad (7)$$
$$V_t = W_V^t SA(z^l) + b_V^t \quad (8)$$

where $W_Q^t, W_K^t, W_V^t \in \mathbb{R}^{D \times D}$ and $b_Q^t, b_K^t, b_V^t \in \mathbb{R}^D$ are temporal attention weights and biases. Temporal attention follows a similar mechanism:

$$TA(SA(z^l)) = softmax\left(\frac{Q_t K_t^T}{\sqrt{d_h}}\right) V_t. \quad (9)$$

Crucially, temporal attention reshapes feature tensors to $[(B \times N) \times T \times D]$, allowing each spatial position to learn long-term dependencies across $T$ time steps.

We introduce a feature attention (FA) module to effectively process the rich grid feature information in RTS tasks, which significantly differs from the RGB values in video pixel features. The FA module processes features after spatiotemporal dependency modeling:

$$Q_f = W_Q^f TA(SA(z^l)) + b_Q^f \quad (10)$$
$$K_f = W_K^f TA(SA(z^l)) + b_K^f \quad (11)$$
$$V_f = W_V^f TA(SA(z^l)) + b_V^f. \quad (12)$$

where $W_Q^f, W_K^f, W_V^f \in \mathbb{R}^{d' \times d'}$ and $b_Q^f, b_K^f, b_V^f \in \mathbb{R}^{d'}$ are feature attention parameters, with $d' = \frac{D}{C} = 31$ representing per-feature channel dimension. Feature attention computation is expressed as:

$$FA(TA(SA(z^l))) = softmax\left(\frac{Q_f K_f^T}{\sqrt{d'}}\right) V_f. \quad (13)$$

At this stage, feature representations are reshaped to $[(B \times T \times N) \times C \times d']$, enabling the model to learn associations between different features such as entity types, health values, and resources.

For deep feature extraction, we combine the three attention modules sequentially to form basic Transformer blocks, constructing a deep network architecture with L=12 layers. The computation process for each Transformer block can be represented as:

$$z^{l+1} = LN\left(FA(TA(SA(z^l) + z^l))\right) \quad (14)$$

where LN denotes Layer Normalization, used to normalize feature distributions and stabilize training. The addition symbol represents residual connections, mitigating gradient vanishing in deep networks. Specifically, the output $z^{l+1} \in \mathbb{R}^{B \times (T \times N+1) \times D}$ of layer $l$ inherits all feature information from layer l-1 while incorporating new feature representations. This cascaded multi-layer design enables the model to construct hierarchical representations from tactical to strategic features: shallow layers focus on local tactical features (such as unit combinations and resource control), while deeper layers integrate global information for strategic understanding.

After L layers of feature extraction, the model performs final situation assessment through nonlinear transformation of the special token vector:

$$y = \sigma(W_2 \varphi(W_1 z_{cls}^L + b_1) + b_2) \quad (15)$$

where $z_{cls}^L \in \mathbb{R}^D$ represents the feature representation of the final layer token vector, $W_1 \in \mathbb{R}^{D \times 4D}$ and $W_2 \in \mathbb{R}^{4D}$ are weight matrices of two fully connected layers, $b_1 \in \mathbb{R}^{4D}$ and $b_2 \in \mathbb{R}$ are corresponding bias vectors, $\varphi$ denotes the GELU activation function, and $\sigma$ is the sigmoid activation function. The output $y \in [0,1]$ represents the predicted probability of victory for one side. This design enables the model to compress complex multi-dimensional feature information into a single situation assessment score.

$$Loss = -\frac{1}{M}\sum_{i=1}^{M}(y_i \log(p_i) + (1-y_i)\log(1-p_i)) \quad (16)$$

The model employs binary cross-entropy as the loss function to measure the discrepancy between predictions and ground truth labels:

where $M$ denotes batch size, $y_i \in \{0,1\}$ is the true winning faction label for the $i$-th sample, and $p_i$ is the corresponding model prediction probability. For optimization, we employ the AdamW optimizer with weight decay coefficient set to 0.01, initial learning rate of 1e-4, momentum decay coefficients $\beta_1 = 0.9$ and $\beta_2 = 0.999$, and numerical stability constant $\varepsilon = 1e-8$. Training parameters for different depths of tri-dimensional attention Transformer (TSTF Transformer) and traditional Timesformer are shown in TABLE II.

TABLE II
INPUT FEATURE PLANES FOR MULTI-SCALE CONVOLUTIONAL NEURAL NETWORK

| Model | De | Number | Mode | Batch | Initial |
|---|---|---|---|---|---|

|  | pth | of Parameters | l Size ( MB) | Size | Learning Rate |
|---|---|---|---|---|---|
| Timesformer | 12 | 5,542,146 | 21.14 | 2 | 0.0001 |
| TSTF Transformer-6 | 6 | 3,565,314 | 13.60 | 1 | 0.0001 |
| TSTF Transformer-8 | 8 | 4,750,082 | 18.12 | 1 | 0.0001 |

Although the channel attention mechanism in the feature dimension increases the parameter count per layer, its introduction enhances the network's feature utilization efficiency, achieving superior performance with fewer layers (e.g., TSTF Transformer-8 with depth 8 outperforms Timesformer with depth 12), resulting in a more compact model design with fewer total parameters and smaller model size.

Considering the complexity and diversity of RTS game situation data, we partition the dataset into training, validation, and test sets in a 10:5:2.5 ratio. This partitioning ensures sufficient training samples while maintaining adequate validation and test samples for evaluating model generalization.

We conducted balanced optimization of key hyperparameters. The number of attention heads ($h = 5$) was selected by balancing feature extraction capability and computational efficiency: too few heads limit the model's ability to capture multi-perspective features, while too many heads lead to excessive computational overhead. The dimension per attention head ($d_h = 31$) was determined based on input feature complexity and computational resource availability.

In the feature attention module, dividing the $D$-dimensional feature space into $C$ subspaces, each with dimension $d' = 31$, significantly reduces attention computation complexity. This design maintains sufficient feature interaction while avoiding excessive computational resource consumption.

Experimental results demonstrate advantages of this tri-dimensional divide-and-conquer attention mechanism design in multiple aspects: first, compared to traditional Timesformer, it achieves better feature utilization efficiency; second, the decomposed attention mechanism enables more precise capture of temporal, spatial, and feature dimension interactions in RTS games; finally, the model demonstrates stable situation assessment capabilities across different map sizes and game scenarios, validating the effectiveness and robustness of this architectural design.

The neural network training environment consisted of an Intel i7 14700K processor, NVIDIA RTX A5000 GPU (24GB), and 64GB RAM, utilizing PyTorch 2.2.0+cu121 (Python 3.10.14).

The training processes for tri-dimensional attention Transformer (depths=6 and 8) and traditional Timesformer are shown in Figure 2. All three models achieved convergence at approximately 10 epochs, with loss values stabilizing.

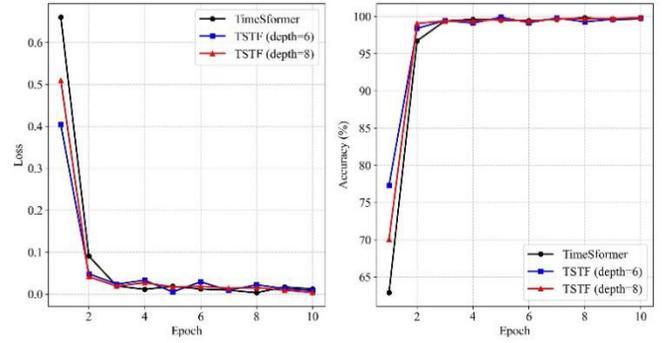

**Figure 2.** Training processes of TSTF Transformer at different depths and traditional Timesformer. The graph demonstrates convergence patterns and accuracy improvements across training epochs.

Comparative analysis reveals that the tri-dimensional attention Transformer models exhibit faster convergence and more rapid accuracy improvement during training, particularly in the first two epochs.

The three neural network architectures—TimeSformer and TSTF Transformer at two different depths (depth=6 and depth=8)—demonstrated distinct training dynamics across 10 epochs. In the initial training phase, TimeSformer showed relatively higher loss (0.66) and lower accuracy (62.89%) compared to both TSTF variants, while TSTF (depth=6) exhibited superior initial performance (loss: 0.4036, accuracy: 77.28%).

During the first two epochs, all three models underwent rapid convergence, with TimeSformer achieving the most significant performance improvement (loss reduction from 0.66 to 0.0906). After epoch 2, the models displayed similar convergence patterns, maintaining stable high accuracy (>99%) and low loss values (<0.02). Notably, TSTF Transformer (depth=8) achieved the lowest loss (0.0037) and highest accuracy (99.83%) at epoch 10, demonstrating that increased network depth indeed enhanced model performance.

*C. Traditional Situation Assessment Models*

To conduct comparative analysis between our neural network approach and conventional methods, we employed two traditional state assessment methods: the built-in Simple evaluation (Simple eval.) and Lanchester evaluation (Lanchester eval.) from microRTS. Both methods utilize weighted scoring systems to assess game states and predict match outcomes.

The Simple eval method employs linear combination of features to calculate player strength:

$$E_p = W_{res}R_p + W_{work}\sum_{u \in W_p} R_u + W_{unit}\sum_{u \in p} \frac{C_u HP_u}{MaxHP_u} \quad (17)$$

where $E_p$ represents the total strength score of player $p$, $R_p$ denotes accumulated resources, $W_p$ represents the worker unit set, $R_u$ indicates resources carried by each worker, $C_u$ represents unit cost coefficient, and $\frac{HP_u}{MaxHP_u}$ reflects the current health ratio of each unit. Weights $W_{res}, W_{work}, W_{unit}$ are built-in microRTS parameters.

The Lanchester eval method incorporates Lanchester's combat laws, considering force concentration effects in military confrontations:

$$E'_p = W_{res}R_p + W_{work}\sum_{u \in W_p} R_u + W_{base}\frac{HP_{base}}{MaxHP_{base}} + W_{barracks}\frac{HP_{barracks}}{MaxHP_{barracks}} + \alpha_u \sum_{u \in p}\frac{HP_u}{MaxHP_u}N_p^{(o-1)} \quad (18)$$

where $\alpha_u$ represents unit-specific strength coefficients and $N_p$ denotes total combat units. The strength coefficients, weights, and exponent 0.7 are built-in microRTS parameters optimized for force concentration effects across various combat scenarios. Both methods process game states represented as 16×16×5 feature matrices, encoding entity types (7 categories), health values (0-10), faction affiliation (1-2), neutral resources (0-25), and accumulated resources (0-25).

For outcome prediction, we calculate $E_1 - E_2$ for Simple eval or $E'_1 - E'_2$ for Lanchester eval, where positive values predict player 1's victory. Comparison with traditional methods helps quantify the advantages of our proposed neural network architecture in capturing complex nonlinear relationships within RTS game states.

*D. Model Evaluation Metrics*

To objectively evaluate the performance of the tri-dimensional attention Transformer model, we employed four standard metrics enabling comprehensive assessment across all dimensions of predictive performance.

Accuracy is defined as the ratio of correct predictions to total predictions:

$$\text{Accuracy} = \frac{\text{Number of Correct Predictions}}{\text{Total Number of Predictions}}. \quad (19)$$

This provides a basic measure of overall model performance, ranging from 0 to 1, where 1 indicates perfect accuracy and 0 indicates complete inaccuracy.

Precision evaluates the accuracy of positive predictions and assesses false positive occurrences:

$$\text{Precision} = \frac{\text{True Positives}}{\text{True Positives} + \text{False Positives}}. \quad (20)$$

This metric is particularly valuable when false positives carry high costs. Like accuracy, precision ranges from 0 to 1, where 1 indicates perfect precision (no false positives) and 0 indicates all positive predictions are incorrect.

Recall, also known as sensitivity, measures the model's ability to detect all actual positive cases:

$$\text{Recall} = \frac{\text{True Positives}}{\text{True Positives} + \text{False Negatives}}. \quad (21)$$

Values range from 0 to 1, where 1 indicates all true positives are correctly identified and 0 indicates no true positives are detected.

F1 Score represents the harmonic mean of precision and recall:

$$\text{Recall} = \frac{\text{True Positives}}{\text{True Positives} + \text{False Negatives}}. \quad (22)$$

This provides balance when both false positives and false negatives carry significant, equal costs. The score ranges from 0 to 1, with 1 indicating perfect precision and recall.

Overall Performance (OP): To reflect comprehensive performance, we introduce the OP index:

$$\text{OP} = \text{Accuracy} + \text{Precision} + \text{Recall} + \text{F1 Score}. \quad (23)$$

Higher OP scores indicate better performance across all key aspects of binary classification. This composite metric provides a holistic view of model performance, reflecting balance among individual metrics. Notably, "optimal" OP refers to best overall performance rather than necessarily achieving the highest score in each individual metric.

## III. RESULTS

To investigate the impact of network depth on situation assessment performance, we systematically analyzed the performance of Transformer architectures with different depths in battlefield situation assessment tasks. Specifically, we designed tri-dimensional attention Transformers (TSTF Transformer) with 6 and 8 layers and compared them with a 12-layer TimesFormer, aiming to validate the relationship between model depth and assessment performance while exploring whether comparable or superior performance could be achieved with fewer layers through optimized architectural design.

Model depth (number of Transformer layers) significantly influences performance, primarily manifesting in feature extraction capability, computational efficiency, and convergence speed. Therefore, we tested TSTF Transformers of different depths (depth=6, depth=8) and compared them with traditional assessment methods to comprehensively evaluate the impact of model depth on situation assessment performance.

First, Figure 3 shows the comparison of situation assessment results between Timesformer (depth=12) and TSTF Transformer (depth=6). This comparison not only reflects performance differences between models of different depths but also demonstrates the advantages of our proposed tri-dimensional attention mechanism in maintaining high performance while reducing model complexity.

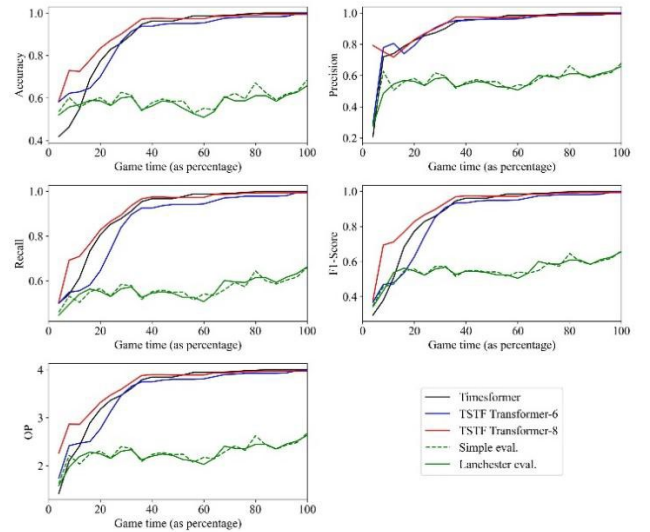

**Figure 3.** Comparison of situation assessment results. Performance comparison between Timesformer (depth=12), TSTF Transformer (depth=6), TSTF Transformer (depth=8), mi-croRTS built-in evaluation function (Simple eval.), and Lanchester model (Lanchester eval.).

Our systematic performance evaluation of five neural network architectures in battlefield situation assessment revealed significant performance differences. Analysis shows that TSTF Transformer-8 demonstrates distinct advantages and limitations in convergence speed, early-stage performance, and stability.

In early-stage performance (around 4% game progress), TSTF Transformer-8 shows significant advantages, achieving 0.587 accuracy compared to TimesFormer's 0.418 and TSTF Transformer-6's 0.582. This early advantage stems from TSTF series' improved feature extraction mechanism, enabling rapid situation awareness establishment in early combat. During mid-game (20-40% progress), the three Transformer architectures display distinct performance curves. TSTF Transformer-8 exhibits the fastest convergence, reaching 0.833 accuracy at 20% progress while maintaining balanced precision (0.829) and recall (0.827), achieving 0.976 accuracy at 40% progress. TimesFormer, despite initial slowness, accelerates during this phase, improving accuracy from 0.773 to 0.962. TSTF Transformer-6 shows stable but relatively slower improvement, with accuracy increasing from 0.782 to 0.938. All three Transformer architectures maintain high performance in late stages (60-100%), while traditional evaluation methods show unstable performance throughout, with accuracy fluctuating between 0.5-0.7, revealing their limitations in complex battlefield environments.

Comprehensive performance evaluation (OP metric) indicates TSTF Transformer-8 maintains the most balanced performance throughout the game. Particularly in the crucial first 40% of game progress, its OP metrics consistently outperform other models, reflecting its practical value in real battlefield applications. From a stability perspective, TSTF Transformer-8 shows significant advantages in early stages (0-40%), with its OP standard deviation (0.947) being 48.6% lower than TimesFormer (1.842) and 24.4% lower than TSTF Transformer-6 (1.253). Traditional evaluation methods' low standard deviations (Simple eval: 0.324, Lanchester eval: 0.298) reflect consistently low performance rather than true performance advantages.

In later stages (40-100%), all Transformer models show significantly reduced standard deviations, indicating stable performance states. TSTF Transformer-8 again performs optimally, with standard deviation of only 0.114, 38.7% lower than TimesFormer (0.186) and 31.7% lower than TSTF Transformer-6 (0.167). Traditional methods show higher standard deviations (Simple eval: 0.283, Lanchester eval: 0.271), indicating performance instability.

TSTF Transformer-8 demonstrates optimal stability across both phases, particularly during the crucial early stage, where its lower standard deviation indicates that increasing input data throughout game progression leads to more reliable performance improvements. While traditional evaluation methods show small early-stage standard deviations, this "stability" actually reflects their limited performance improvement potential.

To visually demonstrate model performance, we selected and visualized two match samples: one with red player 1 victory (Figure 4) and another with blue player 2 victory (Figure 5). Battlefield situation evolution is represented through environmental state abstract feature matrices extracted at 40-step intervals, using geometric symbols (◯:base, ⬢:barracks, ◈:resource point, ○:worker, △:light unit, □:heavy unit, ◇:ranged unit) combined with color coding (red:player 1, blue:player 2, green:neutral resources). Unit health values are numerically annotated. During situation assessment, models assign scores to both players (P1 for player 1, P2 for player 2; scores represent weighted values in traditional methods and victory probabilities in neural network models), with victory predicted based on score comparison (P1>P2 indicates player 1 victory, and vice versa).

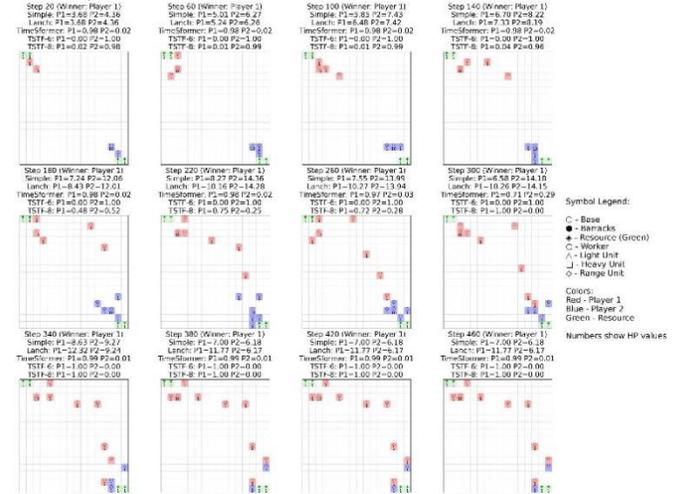

**Figure 4.** Visualization of game progression for red player victory sample. Titles from top to bottom: Step number and labeled winner, Simple eval. prediction, Lanchester eval. prediction, Timesformer prediction, TSTF transformer-6 prediction, TSTF transformer-8 prediction.

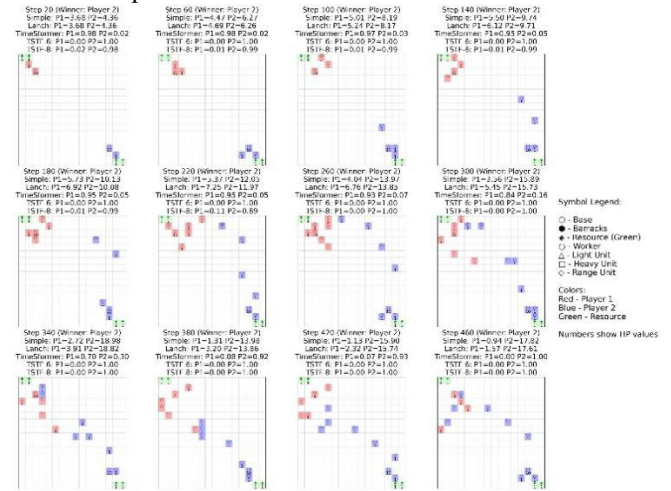

**Figure 5.** Visualization of game progression for blue player victory sample. Titles from top to bottom: Step number and labeled winner, Simple eval. prediction, Lanchester eval. prediction, Timesformer prediction, TSTF transformer-6 prediction, TSTF transformer-8 prediction.

Analysis of these two representative samples reveals that TSTF Transformer-8 demonstrates significant prediction stability throughout the entire match process. Notably, during the mid-game phase (40%-60% game progress, approximately 200-300 steps) characterized by intense resource competition and dramatic battlefield changes, TSTF Transformer-8 maintains stable prediction output while traditional evaluation methods and TimesFormer exhibit significant fluctuations in victory probability predictions. This stability not only validates the model's robustness but also highlights its reliability in complex adversarial environments.

## IV. Discussion

The experimental results demonstrate several significant findings regarding the effectiveness of our proposed TSTF Transformer architecture in RTS game situation assessment. Here we analyze these findings from multiple perspectives and discuss their broader implications.

The superior performance of TSTF Transformer-8 can be attributed to several key technical innovations:

1. Integrated Feature Processing: The incorporation of feature attention mechanisms alongside spatiotemporal attention enables more comprehensive state representation. Unlike traditional Timesformer architectures that primarily process spatial and temporal dimensions, our model's dedicated feature attention module effectively captures the complex interactions between different game state features (unit types, health values, resources, etc.).

2. Efficient Architecture Design: Despite having fewer layers than the baseline Timesformer (8 vs 12), TSTF Transformer-8 achieves superior performance while requiring fewer parameters (4.75M vs 5.54M). This suggests that our tri-dimensional attention mechanism more efficiently processes the structured information present in RTS game states.

3. Superior Performance with Limited Input: The model's ability to achieve 58.7% accuracy with only 4% of game progression demonstrates remarkable effectiveness in early situation assessment. This represents a significant improvement over all other models including Timesformer's 41.8% accuracy at the same stage and highlights the architecture's capability to extract meaningful patterns from limited information.

4. Stability and Consistency: The low performance variation (standard deviation 0.114) during mid-game stages indicates robust and reliable assessment capabilities, crucial for practical applications in strategic decision-making.

This study explores the application of transformer architectures to RTS game state evaluation through decomposing attention mechanisms into specialized components (spatial, temporal, and feature). The experimental results indicate that TSTF Transformer-8 achieves reliable early-game assessment capabilities while maintaining computational efficiency through reduced parameter count. This performance advantage is particularly valuable in RTS games' opening phases, where traditional evaluation methods often struggle to provide dependable assessments.

The model's efficient architecture and ability to process multi-dimensional feature information suggest potential practical applications. In RTS contexts, the reduced parameter count enables real-time processing under computational constraints. Beyond RTS games, similar principles might benefit other domains requiring concurrent analysis of spatial, temporal, and feature relationships, such as autonomous systems or strategic planning.

However, several challenges remain. While more efficient than Timesformer, the tri-dimensional attention mechanism still incurs notable computational overhead. Future research could explore optimization techniques and pruning methods to further reduce computational requirements, as well as investigate the adaptation of these concepts across different applications.

## V. Conclusion

Experimental results demonstrate a clear performance hierarchy: in terms of accuracy, TSTF Transformer-8 > Timesformer > TSTF Transformer-6; in terms of model complexity, Timesformer > TSTF Transformer-8 > TSTF Transformer-6. While the shallower TSTF Transformer (depth=6) outperforms Timesformer (depth=12) during low-information phases (approximately first 10% of game progress), its performance gradually deteriorates relative to Timesformer as information density increases with game progression. TSTF Transformer-8 achieves an optimal balance between accuracy and model complexity, surpassing traditional Timesformer in model efficiency while maintaining superior accuracy throughout the game progression. In contrast, both Timesformer and TSTF Transformer-6 perform comparably or slightly worse than traditional models (Simple eval. and Lanchester eval.) during early stages when input information is limited.